\documentclass[sigconf, screen]{acmart}
\usepackage[rightcaption]{sidecap}
\usepackage{latexsym}
\usepackage{amsmath}
\PassOptionsToPackage{table}{xcolor}
\usepackage{hyphenat}
\usepackage{url}
\usepackage{xspace}
\usepackage{amsmath}
\usepackage{mathtools}
\usepackage{amsfonts}
\usepackage{algorithm}
\usepackage[noend]{algpseudocode}
\usepackage{chngcntr}
\usepackage{caption}
\usepackage{subcaption}
\usepackage[show]{notes-alt}
\usepackage[table]{xcolor}
\counterwithout{algorithm}{section}
\usepackage{balance}
\usepackage{enumitem}
\usepackage{soul}
\usepackage{tikz}
\usepackage{tabularx}
\usetikzlibrary{fit,positioning,calc,shapes}
\usepackage{pgfplots}
\pgfplotsset{compat=1.13}
\usepgfplotslibrary{colorbrewer}
\usepackage{multirow}
\usepackage{breqn}
\usepackage{caption}
\usepackage{stfloats}
\usepackage{booktabs}
\usepackage{subcaption}
\usepackage{stmaryrd}

\DeclarePairedDelimiterX{\norm}[1]{\lVert}{\rVert}{#1}

\newcommand{\approach}{\textsc{FaxPlainAC}}
\newcommand{\expred}{\textsc{ExPred}}

\copyrightyear{2021}
\acmYear{2021}
\setcopyright{acmcopyright}\acmConference[CIKM '21]{Proceedings of the 30th ACM
International Conference on Information and Knowledge Management}{November
1--5, 2021}{Virtual Event, QLD, Australia}
\acmBooktitle{Proceedings of the 30th ACM International Conference on Information
and Knowledge Management (CIKM '21), November 1--5, 2021, Virtual Event, QLD,
Australia}
\acmPrice{15.00}
\acmDOI{10.1145/3459637.3481985}
\acmISBN{978-1-4503-8446-9/21/11}

\begin{document}
\title{FaxPlainAC: A Fact-Checking Tool Based on EXPLAINable Models with HumAn Correction in the Loop}

\author{Zijian Zhang}
\email{zzhang@l3s.de}
\orcid{0000-0001-9000-4678}
\affiliation{
    \institution{L3S Research Center}
    \streetaddress{Appel Str. 9a}
    \city{Hannover}
    \state{Niedersachsen}
    \country{Germany}
    \postcode{30167}
}

\author{Koustav Rudra}
\email{rudra@l3s.de}
\affiliation{
    \institution{L3S Research Center}
    \streetaddress{Appel Str. 9a}
    \city{Hannover}
    \state{Niedersachsen}
    \country{Germany}
    \postcode{30167}
}

\author{Avishek Anand}
\email{anand@l3s.de}
\affiliation{
    \institution{L3S Research Center}
    \streetaddress{Appel Str. 4}
    \city{Hannover}
    \state{Niedersachsen}
    \country{Germany}
    \postcode{30167}
}

\begin{abstract}
Fact-checking on the Web has become the main mechanism through which we detect the credibility of the news or information. 
Existing fact-checkers verify the authenticity of the information (support or refute the claim) based on secondary sources of information. 
However, existing approaches do not consider the problem of model updates due to constantly increasing training data due to user feedback. 
It is therefore important to conduct user studies to correct models' inference biases and improve the model in a life-long learning manner in the future according to the user feedback.
In this paper, we present \approach{}, a tool that gathers user feedback on the output of explainable fact-checking models. \approach{} outputs both the model decision, i.e., whether the input fact is true or not, along with the supporting/refuting evidence considered by the model. Additionally, \approach{} allows for accepting user feedback both on the prediction and explanation.
Developed in Python, \approach{} is designed as a modular and easily deployable tool. It can be integrated with other downstream tasks and allowing for fact-checking human annotation gathering and life-long learning.

\end{abstract}




\maketitle
\section{Introduction}
\label{sec:intro}
Massive amounts of information is produced daily on the Web and social media~\cite{holzmann2016dawn}. However, not all of it is credible, and it is usually beyond the human scope to verify each one manually. 
Sometimes, non-credible information inadvertently gets propagated by even popular personalities and manages to reach a large audience. 
In recent times, researchers proposed different fact-checking algorithms~\cite{kotonya-toni-2020-explainable,zhang2021expred} and try to automatically flag inaccurate content. 
However, most of these models are obscured in nature and rely on the explicit supervised data. 
Even for the models that indeed predict the correctness of claims,  a general lack of interpretability inhibits general trust over its utility in general.

\begin{figure}[tb]
    \centering
    \includegraphics[width=\columnwidth]{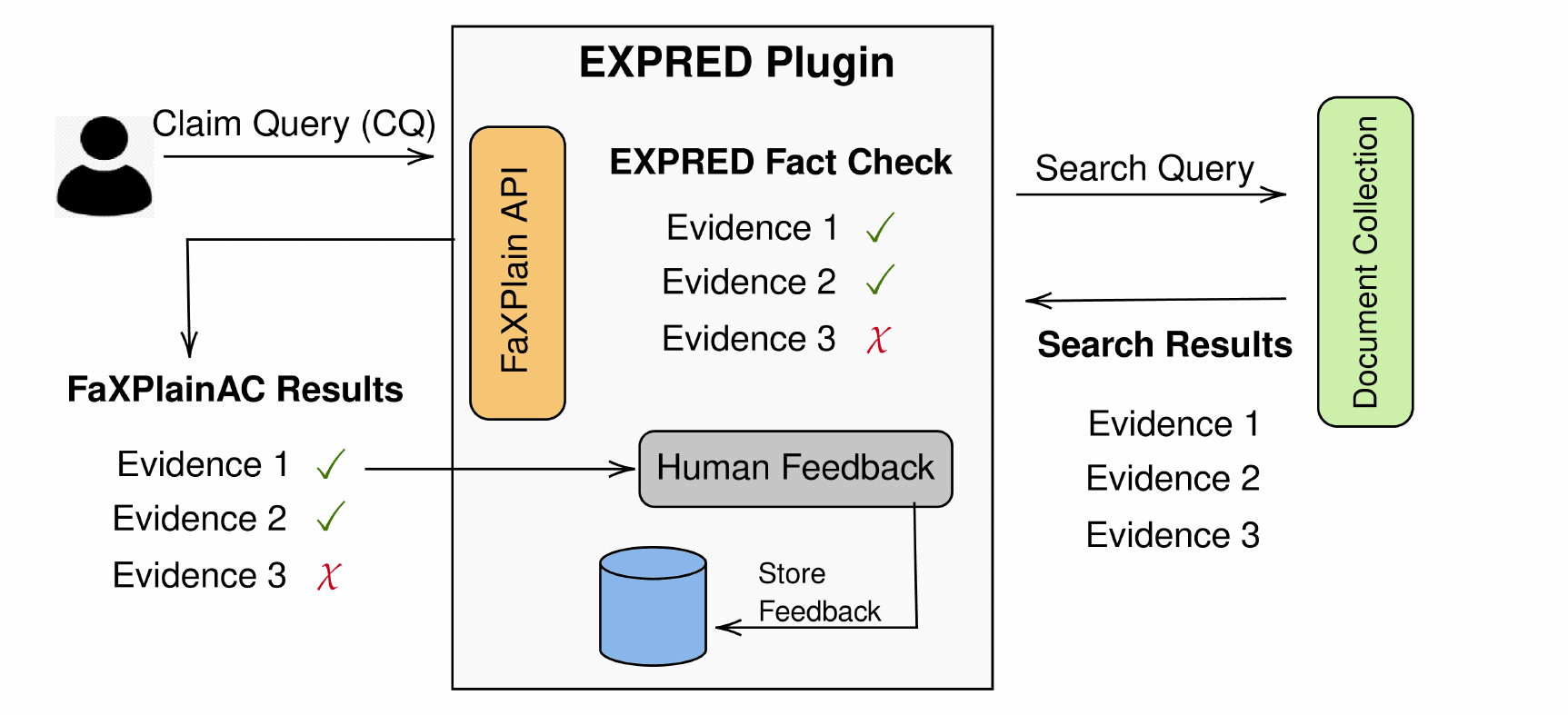}
    \caption{Architecture of \approach.}
    \label{fig:event_seq}
\end{figure}

Towards this, there have been some recent proposals that create models that are \textit{interpretable-by-design approach} that generate an explanation alongwith the actual fact-checking prediction~\cite{lei2016rationalizing,bastings2019interpretable,zhang2021expred}. 
Specifically, these approaches not only perform fact-checking on a piece of information but also output the evidence/reasons behind the model's decision in terms of an extractive piece of text with the end-users. 
Moreover, approaches like \textsf{Quin+}\cite{samarinas2021improving} provides a web front-end to present not only verification of the user's claim but also the evidence from the web supporting the verification. In such approaches, the interpretable-by-design models serve as the engine of the fact-checking system. Such interpretable-by-design models are, unfortunately, usually trained on datasets with limited size like \textsc{Fever}\cite{thorne-etal-2018-fever}. Such limited training data introduces a certain bias towards the used dataset. 
The end-users, therefore, cannot fully trust the model's prediction or interpretation. 
They would also have demand to correct the model's erroneous prediction or explanation.
 A human intervention, fortunately, is easy to scale up by conducting crowdsourcing tasks \cite{zhang2019dissonance}. 

In this paper, we present \approach\footnote{\url{https://faxplain.appspot.com/}}, the first interpretable open-source fact-checking web front-end that gathers user correction on model's false prediction/interpretation. Our framework is  presented in Figure.~\ref{fig:event_seq}. \approach{} first retrieves documents potentially related to the claim from online encyclopedias, followed by fact-checking using the recent interpretable by design approach \expred~\cite{zhang2021expred}. The approach \expred{} justifies (support/refute) the claim regarding each document, together with provides evidence supporting the inference (details in Section~\ref{sec:background}); and at last ask users feedback, or correction by false results, about the model's prediction and interpretation. \approach{} is implemented in Python with the objective to assist users with supporting/refuting evidence along with the decision of the model, together with gathering user annotation in fact-checking task. 
At present, we incorporate \expred{} only into our framework. However, our modular design allows us to incorporate different standalone interpretable fact-checking algorithms into \approach{}, so that the benefits of such proposed algorithms become accessible to different stakeholders such as end-users, professional developers, tech companies, etc. 
For example, one can easily use an interpretable search system~\cite{singh2019exs,singh2020model,fernando2019study}, or use evidence-gathering mechanisms~\cite{singh2016expedition,singh2016discovering,holzmann2016tempas}.
We believe such an application will be immensely helpful in developing new socio-technical systems that rely on automated fact checking.
\section{\expred: Method Background}
\label{sec:background}
This section explains the idea behind \expred~\cite{zhang2021expred} approach and gives examples from the fact-checking domain. \expred{} is a two-phase pipeline interpretable by a design approach that incorporates the task prediction and the rationales. 

In the first phase, it learns to extract explanations/rationales from the input document supervised by both task-specific and rationale-based signals under a multi-task learning (MTL) framework. Both predictions and explanations are learned using a common BERT\cite{devlin2018bert}-based encoder. However, separate decoders are utilized for task and explanation extraction. The MTL criteria are the linear combination of both loss for the prediction and explanation task and represented as follows:
\begin{equation}
 \label{eqn:loss_function}
 \mathcal{L}_{loss} = \mathcal{L}_{task} + \lambda \mathcal{L}_{exp}\mbox{,}
\end{equation}
where $\mathcal{L}_{task}$ is a standard binary cross-entropy loss and is the loss of the auxiliary prediction task. The $\mathcal{L}_{exp}$ is the explanation loss. Because there are far more non-evidence tokens than evidence tokens in the explanation supervision, the explanation loss is a balanced version of explanation loss, i.e.
\begin{equation}
    \mathcal{L}_{exp}=\frac{1}{|t_{0}|}\mathtt{
        BCE}(p_{0}, t_{0}) + \frac{1}{|t_{1}|}\mathtt{
        BCE}(p_{1}, t_{1})\mbox{,}
\end{equation}
where $|t_{0}|$, $|t_1|$ are the count of the non-evidence/evidence tokens, correspondingly, while $\mathtt{BCE}(p_{0}, t_{0})$, $\mathtt{BCE}(p_{1}, t_{1})$ are the cross entropy for all non-evidence/evidence tokens.

\begin{figure}[tb]
\centering
\includegraphics[width=0.9\columnwidth]{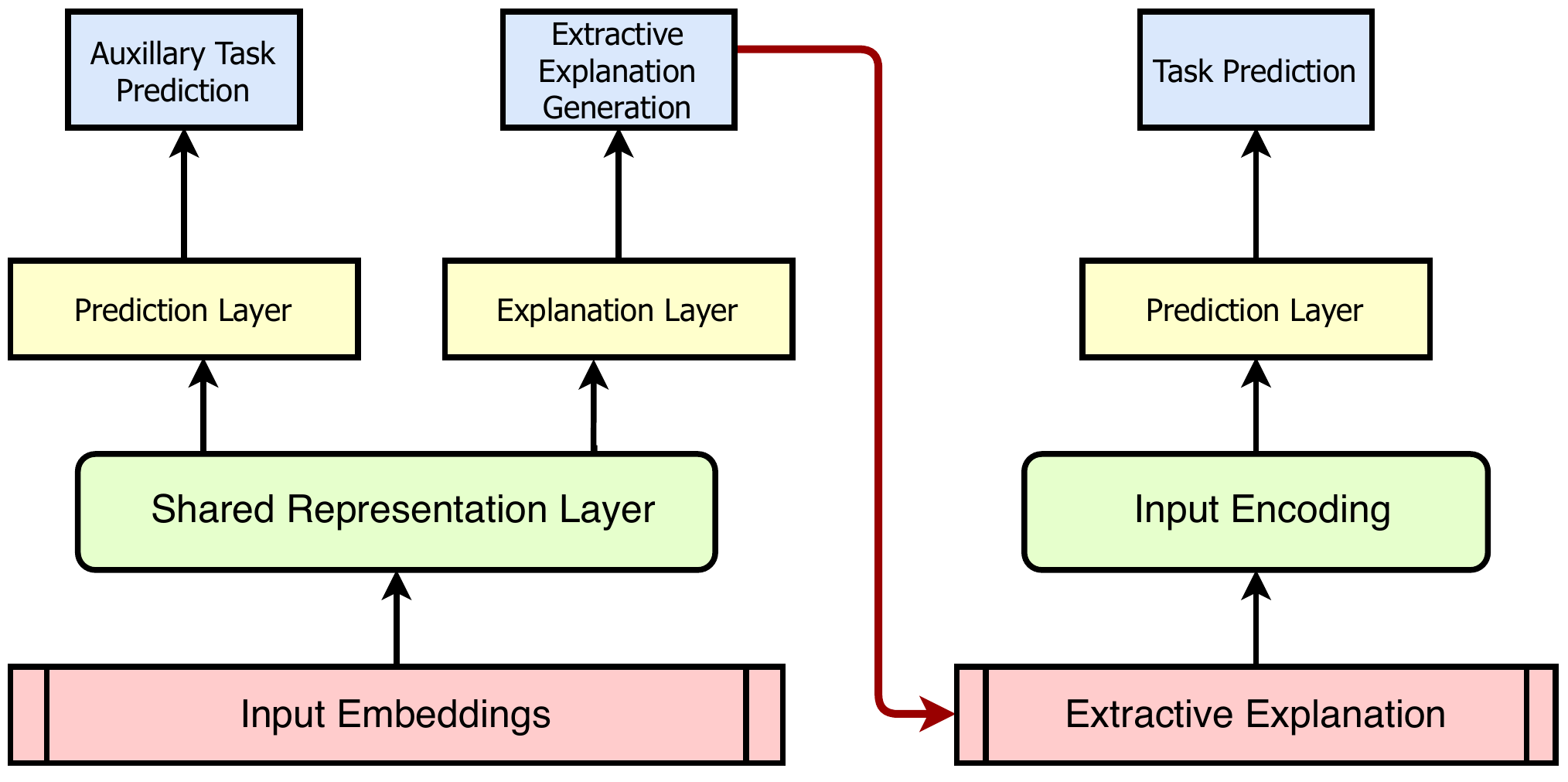}
\caption{\small{Overview of \expred{}.  Explanation generation supervised by \textit{task objectives} and \textit{explanation generation}. Here auxiliary task is the same as the actual task (the Task Prediction on the right). }}\label{fig:mtl-example}
\end{figure}

In the second phase, a separate over-parameterized model is trained for the prediction task only based on the extracted explanations of the first phase. After training the first-phase MTL model, we do inference on the training/validation data once again to emit machine annotated rationales. Then we screen out all such instances, on which the auxiliary prediction branch fails to predict the correct labels. The screening prevents the second-phase model from being polluted by untrustworthy pieces of evidence. Then for the instances left,  all tokens corresponding to evidence identified by the phase 1 model are preserved and all non-evidence tokens are replaced by a wildcard token (here we use period '.'). Then the second-phase classifier is trained on the masked dataset. This scheme is referred to as predict and explaining (first phase) and then predicting again (second phase). Fig.~\ref{fig:mtl-example} describes the overall architecture of \expred{}. 
For example, we want to fact check \textit{Emma Watson was killed in 1990} based on the document \textit{Emma Charlotte Duerre Watson (born 15 April 1990) is a French-British actress, model, and activist. Born in Paris ... previously. Watson appeared in all eight Harry Potter films from 2001 to 2011, earning worldwide fame, critical accolades, and around \$60 million..}. The first sentence \textit{Watson ... (born 15 April 1990) ...}' suffices to refute the claim.

In the initialization phase of \approach{}, an \expred{} model is pre-trained on the FEVER dataset\cite{thorne-etal-2018-fever}. In the inference phase, we concatenate the user input statement and the document following the same manner as presented in \cite{zhang2021expred}, namely \textbf{[CLS] statement [SEP] document}. The model outputs both justification and explanation.

\begin{figure*}[h]
    \centering
        \includegraphics[width=0.75\textwidth]{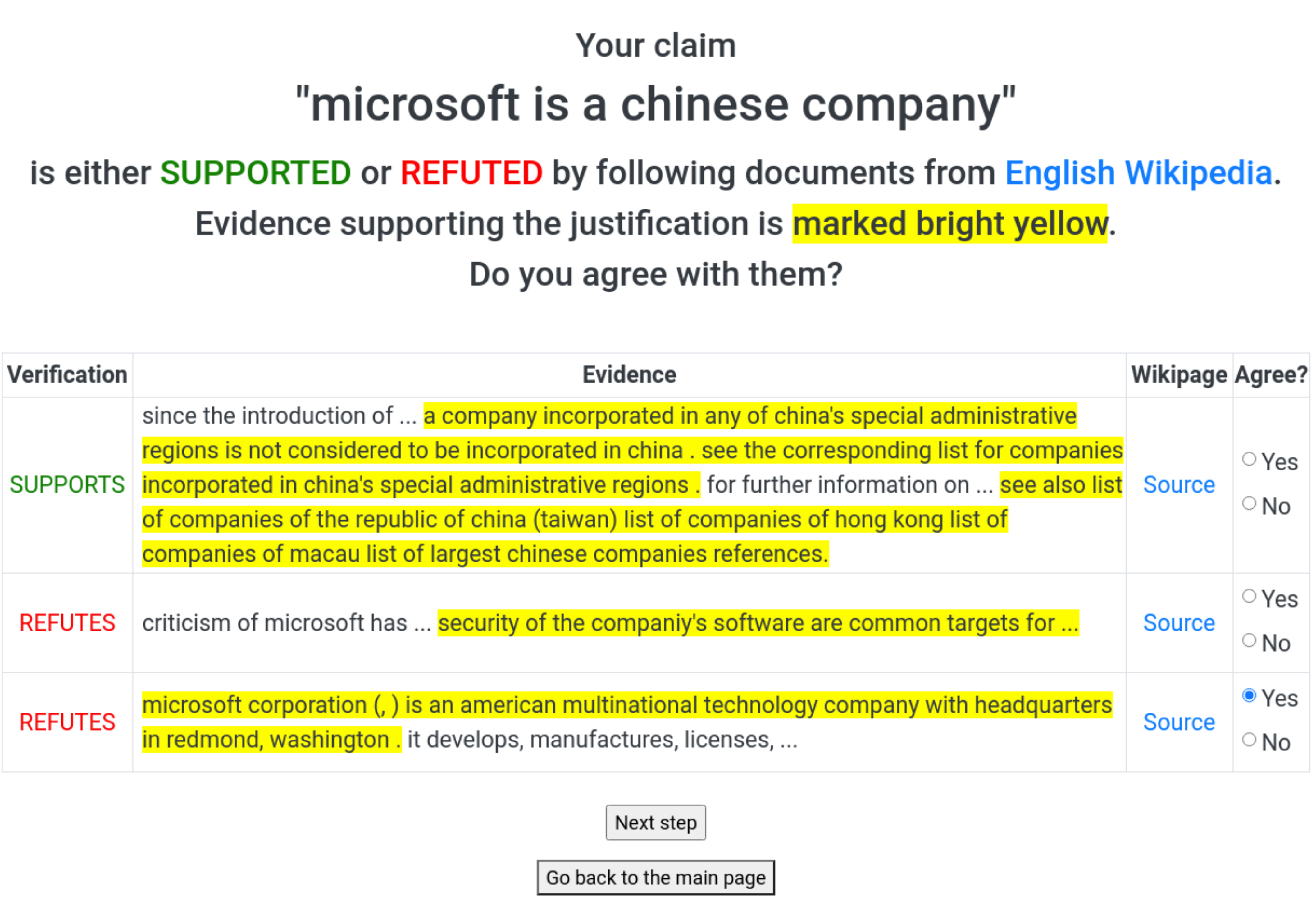}
    \caption{Prediction results}
    \label{fig:predict}
\end{figure*}

\section{\approach{}~as an \expred~ Plugin}
\label{sec:demo}
In this section, we describe the architecture of \approach{} and the training procedure of \expred. The web application part is implemented using Python3 with the Flask\footnote{\url{https://flask.palletsprojects.com/}}~framework. \expred{} model is firstly implemented using Pytorch\footnote{\url{https://pytorch.org/}} 1.8.1 and trained on an Nvidia A100. Then the trained model is loaded through the CPU in the Dockers for inference. Figure~\ref{fig:event_seq} presents the overall architecture of \approach. This decoupled design guarantees our \expred{} model can be interchangeable with other interpretable fact-checking engines. The step-wise usage instruction is as follows:

\subsection{Overview of Search System}
    The \approach{} architecture receives a query from the user. The task of the system is to \textcolor{green}{support} or to \textcolor{red}{refute} the claim based on evidence extracted from documents.
    
\subsubsection{Firing Queries}
    Standard preprocessing steps (stripping, case folding, etc.) are applied to the statement and the \approach{} plugin issues a Google Programmable Search\footnote{\url{https://cse.google.com/cse/}} to retrieve relevant Wikipedia pages for the given claim through Google JSON API\footnote{\url{https://developers.google.com/custom-search/v1/introduction}}. 
    
\subsubsection{Pruning Search Results}
    The top-ranking search results are collected, and relevant content is extracted from those pages with the help of Mediawiki\footnote{\url{https://www.mediawiki.org/wiki/API:Get_the_contents_of_a_page}} RESTful tool. The end users can tell the system how many articles to be returned. Specifically for our running example, we choose the top three articles since they are the most relevant document for this user input.
    
\subsection{Interpretable Fact-checking}
    Next, the \expred{} model is applied over the statement and each of these three documents. The model either \textcolor{green}{SUPPORTS} or \textcolor{red}{REFUTES} the claim, along with providing evidence extracted from documents supporting the prediction. A sample result page is shown in Figure~\ref{fig:predict} for the statement \textit{`Microsoft is a Chinese company'}. Here the first document is misleading for the justification since it states companies in China instead of the company Microsoft. The second seems irrelevant to our statement. And the third result states that Microsoft is \textit{an American multinational company}, which refutes the claim and of which both the prediction and evidence are correct.

\subsection{Collecting User Feedback}
    As mentioned previously, the pre-trained, well-performing model can be erroneous during inference and possibly need to be re-trained on human tutelage: The last column asks users for their feedback. The results agreed by the users (choosing \textbf{Yes} in the column \textbf{Agree?}) are dumped as trustworthy machine-generated evidence, while users being asked for their feedback about the failing results by choosing \textbf{No} and clicking the \textbf{Next step} button. The outcomes of our example corresponds to three different categories of user feedback: 
    \begin{itemize}
        \item the first outcome is wrong because it is misleading for the justification. The user can hence choose \textbf{No} for the column \textbf{Agree?} to report the falsely retrieved document;
        \item the second outcome is irrelevant to the nationality of Microsoft. Again the user can choose \textbf{No} and report the irrelevance;
        \item the third output is correct. The User can simply choose \textbf{Yes} and there is nothing more to do to this one.
    \end{itemize}
    Note that a Wikipedia article is usually too long and might be cumbersome for the users to read-through and later on fully annotate. Also it exceeds the input length limit of the interpretable fact-checking model. Fortunately for the majority of queries, the definition of the related topic is enough for fact-checking. The definition of concepts usually occurs at the beginning of its Wiki page. Hence, we take the first 30 sentences at the beginning of each article by default and display highlighted evidences, together with those non-evidence tokens, if they occurs at the beginning of the article or closely after a piece of evidence. Additionally, a "show more of this document" button is placed for each article. If the user is sure the article is indeed relevant to the query, but the evidence does not present in the selected sentences, the user can acquire the next (e.g.) 30 sentences from the returned article and update the corresponding outcome.
    
\subsection{Token-level User Feedback}
\begin{figure*}[tb]
    \centering
    \includegraphics[width=0.8\textwidth]{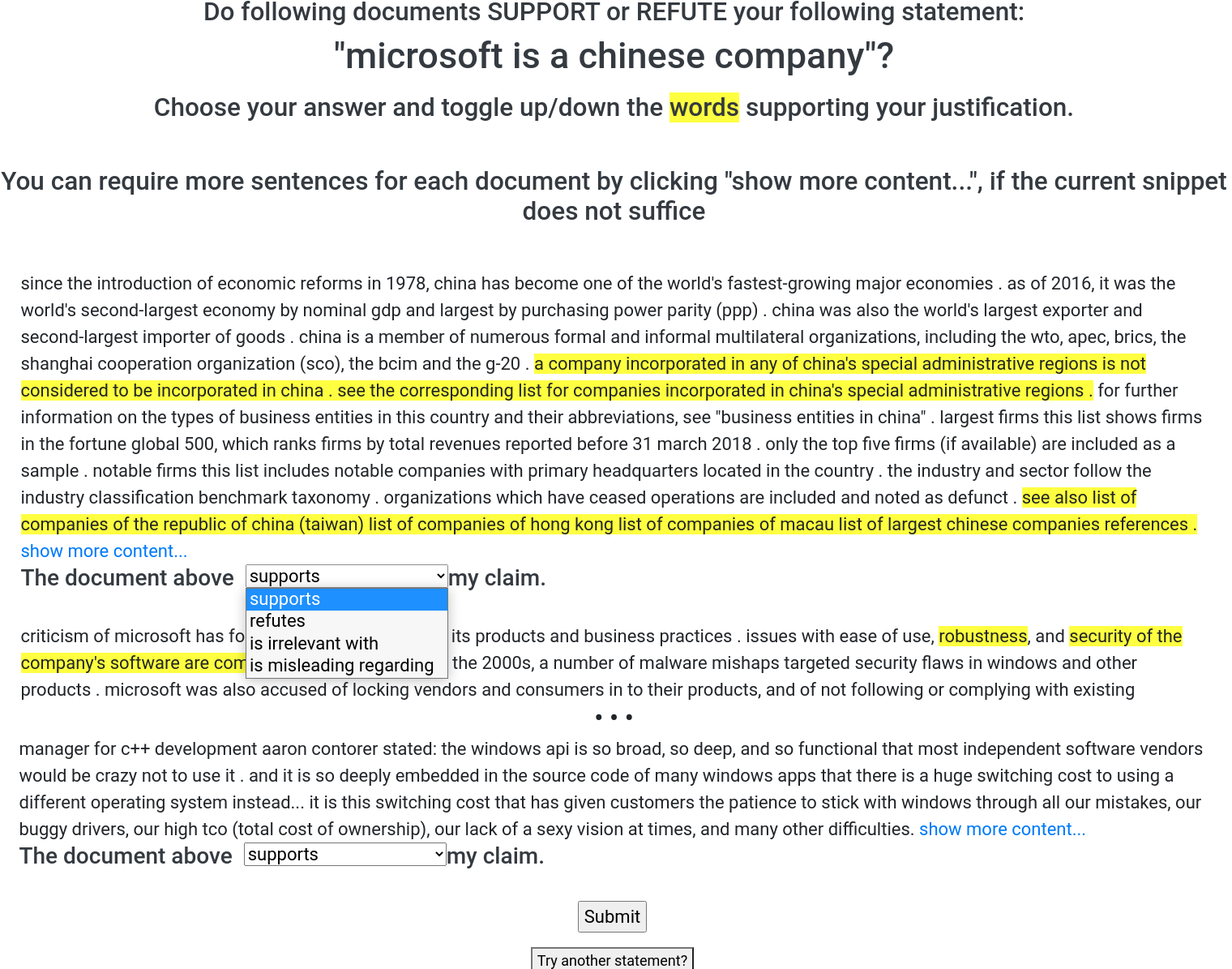}
    \caption{User feedback collection}
    \label{fig:feedback1}
    \vspace{-4mm}
\end{figure*}

    In the final step, we ask users for the following types of feedback on a separate correction page as shown in Figure.~\ref{fig:feedback1}: 
    \begin{itemize}
        \item If the corresponding document either \textcolor{green}{SUPPORTS} or \textcolor{red}{REFUTES} the statement, users are asked to toggle up all evidence tokens supporting the judgment from the document by highlighting them. The machine-annotated rationales are highlighted by default. All highlighted tokens can be toggled down. Multiple selections is also viable through pressing-holding and sweeping the mouse through all evidence tokens.
        \item Users can also choose if the document misleads the justification. For example, the first result in our example incites a \textcolor{green}{SUPPORTS} result since the claim is about companies in China. The user therefore selects \textbf{This document is misleading regarding my claim.}
        \item The document may be irrelevant (like the second one) to the given claim. In the end, such feedback is saved into the user annotated evidence database by clicking the \textbf{Submit} button. Those gathered human annotations are the final output of our \approach{} and can be used to fine-tune the \expred{} model.
    \end{itemize}

From any of the pages mentioned above, the user can go back to the homepage and verify another claim by clicking the '\textbf{Try another statement}~(or \textbf{Go back to the main page})' button. If no results are retrieved from Wikipedia.org, an error page pops out and asks the user to try other statements.

The whole application is at first encapsulated as a docker image and later on deployed on the Gcloud run\footnote{\url{https://cloud.google.com/run}}. The visualization technique and cloud platform can also be replaced and tailored according to the researcher's demand.

\section{Conclusion}
\label{sec:conclu}
In this paper, we present \approach, the first open-source framework for gathering human annotations for interpretable fact-checking systems. 
We leverage the recent interpretable-by-design model \expred{} as the back-end model of our fact checking-explanation-feedback web framework. 
Our decoupled design enables easy integration with other interpretable search systems.
Furthermore, one can switch explainable models because of the modular design of the web front-end and the models. 
The final contribution of our \approach{} is two folds: Firstly, it identifies errors models made on evidence identification or prediction and secondly, it gathers human annotation as correction on the erroneous results. Such human annotations can expand the existing fact-checking datasets and help models to learn from failures.

\vspace{3mm}
\noindent\textbf{Acknowledgement:}
Funding for this project was in part providedby the European Union’s Horizon 2020 research and innovation program under grant agreement No 832921 and No 871042.

\newpage
\bibliographystyle{ACM-Reference-Format}
\bibliography{ref}
\balance

\end{document}